\documentclass[10pt,twocolumn,letterpaper]{article}

\usepackage{wacv}              

\usepackage{graphicx}
\usepackage{amsmath}
\usepackage{amssymb}
\usepackage{booktabs}
\usepackage{times}
\usepackage{epsfig}
\usepackage{pifont}
\usepackage{multirow}
\usepackage{algorithm}
\usepackage{algpseudocode}
\usepackage[table]{xcolor}

\newcommand{\datr}{DATr}
\newcommand{\bestresult}[1]{\text{#1}}
\newcommand{\topresult}[1]{{\color{red}{\textbf{#1}}}}
\newcommand{\secondresult}[1]{{\color{blue}{\textbf{#1}}}}

\usepackage[capitalize]{cleveref}
\crefname{section}{Sec.}{Secs.}
\Crefname{section}{Section}{Sections}
\Crefname{table}{Table}{Tables}
\crefname{table}{Tab.}{Tabs.}

\def\wacvPaperID{155} 
\def\confName{WACV}
\def\confYear{2024}

\begin{document}

\title{Leveraging the Power of Data Augmentation for Transformer-based Tracking}
\author{Jie Zhao\textsuperscript{1}, Johan Edstedt\textsuperscript{2}, Michael Felsberg\textsuperscript{2}, Dong Wang\textsuperscript{1}, Huchuan Lu\textsuperscript{1}\\
\textsuperscript{1}Dalian University of Technology,
\textsuperscript{2}Linköping University\\
{\sl{\small{zj982853200@mail.dlut.edu.cn, \{johan.edstedt,michael.felsberg\}@liu.se,\{wdice,lhchuan\}@dlut.edu.cn}}}
}

\maketitle

\begin{abstract}
   Due to long-distance correlation and powerful pretrained models, transformer-based methods have initiated a breakthrough in visual object tracking performance. Previous works focus on designing effective architectures suited for tracking, but ignore that data augmentation is equally crucial for training a well-performing model. In this paper, we first explore the impact of general data augmentations on transformer-based trackers via systematic experiments, and reveal the limited effectiveness of these common strategies. Motivated by experimental observations, we then propose two data augmentation methods customized for tracking. First, we optimize existing random cropping via a dynamic search radius mechanism and simulation for boundary samples. Second, we propose a token-level feature mixing augmentation strategy, which enables the model against challenges like background interference. Extensive experiments on two transformer-based trackers and six benchmarks demonstrate the effectiveness and data efficiency of our methods, especially under challenging settings, like one-shot tracking and small image resolutions.
\end{abstract}
\vspace{-1em}

\section{Introduction}
\label{sec:intro}
With the development of deep models, many visual object tracking (VOT) works~\cite{siamrpn++,zhang2019deeper,ATOM,chen2021transt,wang2021transformer} focus on designing effective tracking frameworks with modern backbones. Some large-scale tracking datasets with high-quality manual annotations~\cite{lasot,huang2019got10k,trackingnet} are also developed to satisfy these data-driven models. However, a crucial issue is long neglected, that is, appropriate data augmentation is the cheapest strategy to further boost the tracking performance. We notice that most trackers follow similar data augmentation strategies, which are combinations of random cropping and several image transformations, like flip and blur. State-of-the-art (SOTA) transformer-based methods also apply the same pattern as prior works based on convolutional neural networks (CNN). Bhat~\etal~\cite{updt} demonstrated that these general data augmentations (GDA) play an important role on CNN-based trackers. However, considering the substantial difference between CNN and transformer models, and powerful capabilities of transformer models themselves, what is the impact of GDAs on SOTA transformer-based trackers? We think this is a question worth exploring. While it has been demonstrated in several works~\cite{cubuk2018autoaugment,ghiasi2021copypaste,yun2019cutmix,zhang2018mixup,liu2022tokenmix} that well-designed data augmentation is useful for multiple computer vision tasks, few works apply the latest data augmentations or customize suitable approaches for VOT.
\begin{figure}[t]
    \centering
    \includegraphics[width=0.84\linewidth]{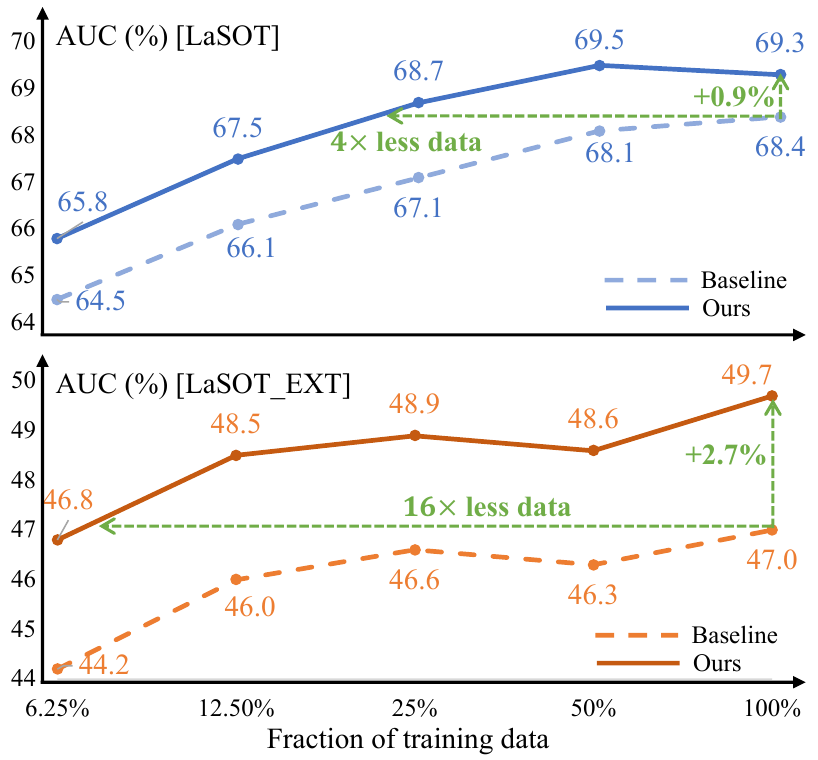}
    \vspace{-0.5em}
    \caption{\textbf{Data-efficiency comparison under different volumes of training data.} Our augmentations are highly effective for low amounts of data, but remarkably also provide large improvements in the large data regime. Results are averaged over 3 random seeds.}
    \vspace{-1.5em}
    \label{fig:data_efficiency}
\end{figure}

\begin{figure*}[t]
    \centering
    \includegraphics[width=\linewidth]{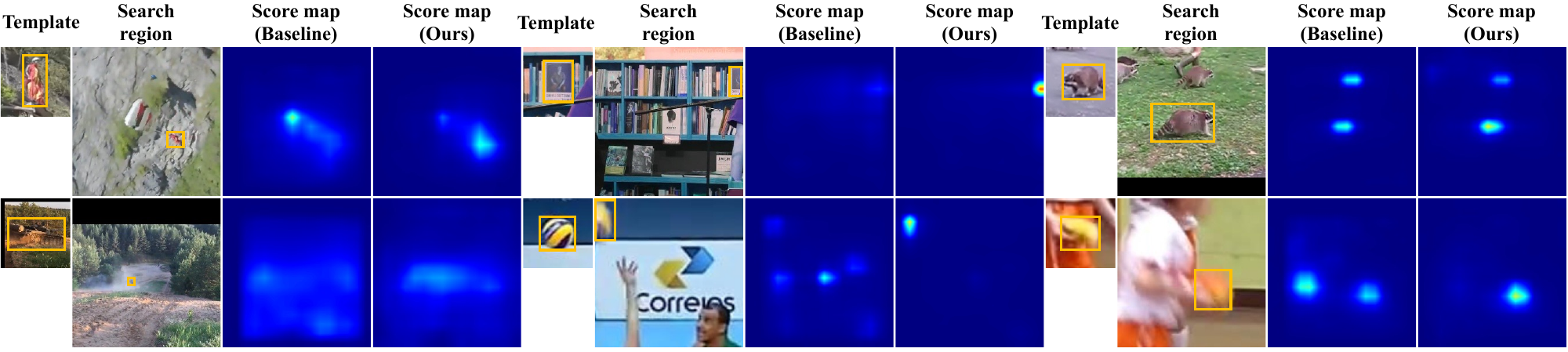}
    \caption{\textbf{Qualitative comparison of score maps on hard cases.} Left: Huge scale variations. Middle: Boundary samples caused by previous inaccurate prediction or fast motion. Right: Interference from the background. Better viewed with zoom-in.}
    \vspace{-1.0em}
    \label{fig:scoremap}
\end{figure*}
In this paper, we perform comprehensive experiments to explore the impact of GDAs on transformer-based trackers, including the pure transformer tracker and the hybrid CNN-Transformer tracker. Different from the conclusion in~\cite{updt}, our experiments imply that most common GDAs have limited effectiveness for these SOTA trackers. We also gain the insight that while models can benefit from increasing jitter for random cropping, large jitters will degrade performance. Moreover, as shown in Fig.~\ref{fig:scoremap}, we find that in addition to the sequence's own challenges, previous inaccurate predictions also cause difficult search patches with huge scale variations (Left) and boundary targets (Middle). Background interference is also challenging for SOTA trackers (Right).

Motivated by these observations, we propose two data augmentation approaches customized for VOT. First, we propose an optimized random cropping (ORC) consisting of a dynamic selection mechanism of search radius factor $\gamma$ and simulation of boundary samples. The former enriches samples from the perspective of context via two-step randomness, which enables the model more robust to scale variations (Fig.~\ref{fig:scoremap} (left)), and furthermore makes the model flexible to different $\gamma$ during inference. The latter helps the model recover fast from failure cases and deal with challenges like fast motion better (Fig.~\ref{fig:scoremap} (middle)). 
Second, we propose a token-level feature mixing augmentation (TFMix). Token features of another object are mixed into the original search features as a distractor. This method makes the model better able to cope with complex background interference (Fig.~\ref{fig:scoremap} (right)). 

Experiments in Sec.~\ref{sec:experiments} demonstrate that our methods not only further boost modern trackers' performance, especially under challenging settings, but also unbind strong association for specific value of $\gamma$ between training and inference. Furthermore, to explore the data efficiency benefit from our methods, we use different volumes of data for model training, \ie~randomly choosing a fraction of sequences from each training dataset. Since we find that reducing the numbers of training sample pairs for settings with small data volumes has little effect on the performance, we follow the same number of sample pairs as the baseline setting (OSTrack256). As shown in Fig.~\ref{fig:data_efficiency}, using only 6.25\% of the data, our methods achieve comparable result on LaSOT\_EXT to the baseline trained with full data. 
  

The main contributions of this work are as follows:
\begin{itemize}
\setlength{\itemsep}{0pt}
\setlength{\parsep}{0pt}
    \item We perform systematic experiments to explore the impact of General Data Augmentations (GDA) on transformer trackers, including the pure transformer tracker and the hybrid CNN-Transformer tracker. Results show GDAs have limited effects on SOTA trackers. 
    \item We propose two \textbf{D}ata \textbf{A}ugmentation methods based on challenges faced by \textbf{Tr}ansformer-based trackers, \datr~for short. They improve trackers from perspectives of adaptability to different scales, flexibility to boundary targets, and robustness to interference.
    \item We apply \datr~to two transformer trackers on six tracking benchmarks, demonstrating the effectiveness and generalization of \datr, especially for sequences with challenges and unseen classes. Experiments on CNN backbones further show the significant generalization effect of our optimized random cropping.
\end{itemize}

\section{Related Work}
\subsection{Visual Object Tracking}
In terms of the types of backbones, tracking methods have gone through three stages of evolution, \ie~traditional approaches~\cite{kalal2011TLD,bolme2010MOSSE,kcf} using hand-crafted features, CNN-based methods~\cite{SiamFC,siamrpn++,zhang2019deeper,dimp}, and transformer-based methods~\cite{chen2021transt,yan2021stark,liu2021swin,ye2022ostrack,cui2022mixformer}. Among them, SiamRPN++~\cite{siamrpn++} and SiamDW~\cite{zhang2019deeper} analyzed the negative effect of large receptive field and padding issue caused by the deep CNN, and investigated proper architectures to make the tracking benefit from very deep CNN. To make up for the locality of CNN, Chen~\cite{chen2021transt}~\etal developed a transformer-based correlation module, which can establish long-distance associations between the template and search region. Recently, several works, \eg~OSTrack~\cite{ye2022ostrack} and MixFormer~\cite{cui2022mixformer}, combined the feature extraction and fusion modules into a whole through a pure transformer architecture, boosting the tracking performance to a new level.

\begin{figure*}[ht]
    \centering
    \includegraphics[width=0.97\textwidth]{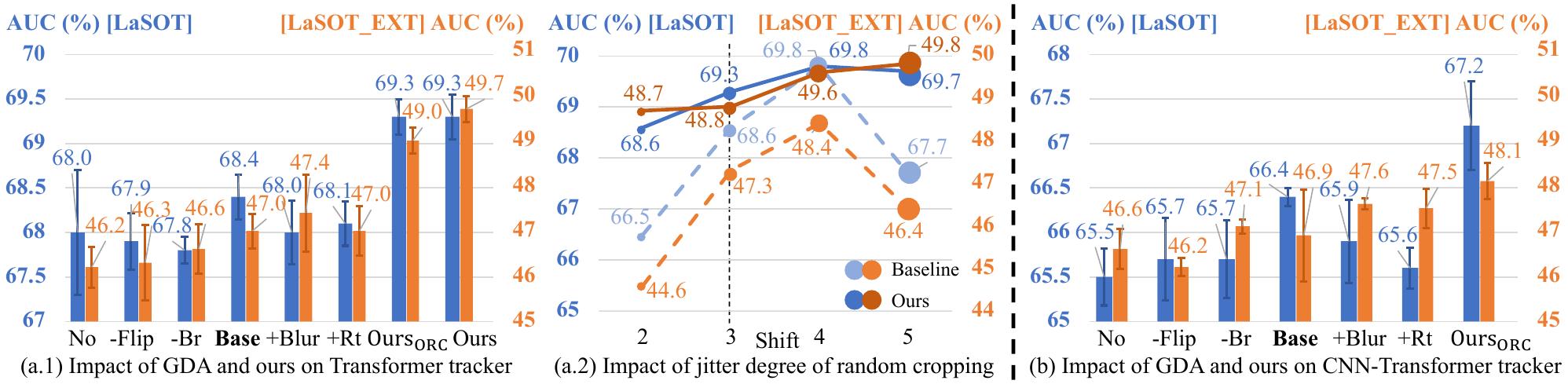}
    \caption{\textbf{Systematic analysis of GDAs, and comparison with ours on LaSOT (blue) and LaSOT\_EXT (orange).} (a.1) and (b) compare the impact of GDA and ours on the pure transformer tracker and the CNN-Transformer tracker, respectively. Results imply a limited effectiveness of GDAs for these SOTA trackers, while ours improve their performance significantly on each benchmark. (a.2) shows the existing random cropping causes model degradation under small and large jitter, while ours are stable for different jitter settings.}
    \label{fig:aug}
    \vspace{-1.em}
\end{figure*}

\subsection{Data Augmentation in Tracking}

Most previous works in tracking focus on designing effective model architectures, or integrating modern backbones into tracking framework. In contrast, despite data augmentation playing a crucial role in the performance of trackers, far less attention has been paid to this topic. 

\noindent \textbf{Augmentations for CNN-based trackers.}
Zhu~\etal~\cite{zhu2018distractor} investigated the important role of diverse training samples. Bhat~\etal~\cite{updt} compared performance gains from GDAs for shallow and deep features and found that deep CNNs particularly benefit from augmentation. Motivated by model robustness to rapid motion, several works~\cite{updt,wang2020cnn,zhu2018distractor} emphasized the impact of blur augmentation.

\noindent \textbf{Augmentations for transformer-based trackers.}
Transformers have been found to exhibit different properties~\cite{naseer2021intriguing} and be more robust to perturbations~\cite{bhojanapalli2021understanding,paul2022vision} than CNNs. Despite this, previous transformer-based trackers~\cite{ye2022ostrack,cui2022mixformer,mayer2022transforming} still use similar data augmentation approaches as for CNNs, and the impact of these augmentations has not been investigated. In contrast to previous works, we systematically investigate the role of GDAs for modern transformer-based trackers (see Sec.~\ref{sec:gda_analysis}). Motivated by experimental observations, we further propose two data augmentation approaches based on challenges faced by modern transformer-based trackers (see Sec.~\ref{sec:datr}).

\subsection{Image Mixing Augmentation}

In the context of computer vision, in addition to the basic geometric transformations (\eg~rotation, scaling, shear, flip), and photometric transformations (\eg~saturation, grayscale, color-jittering), a multitude of augmentations obtain diverse data via mixing different images. For example, MixUp~\cite{zhang2018mixup} blends images pixel by pixel. CutMix~\cite{yun2019cutmix} replaces contents of a random area with a patch cropped from another image. Customized for transformer models, TokenMix~\cite{liu2022tokenmix} mixes images in units of tokens. The effectiveness of these mixing methods has been demonstrated in many tasks, such as object detection~\cite{wang2021pyramid,dwibedi2017cut}, instance segmentation~\cite{ghiasi2021copypaste}, and video classification~\cite{yun2020videomix}, but few works integrate this type of methods into VOT. To the best of our knowledge, the only work to apply a similar strategy to tracking is~\cite{li2023self}, which performs crop-transform-paste operations on images for self-supervised tracking. 
Unlike this, we propose a token-level feature mixing strategy to simulate background interference.

\section{Analysis of General Data Augmentation}
\label{sec:gda_analysis}

General data augmentations (GDA) are ubiquitously used in tracking. As shown in Tab.~\ref{tab:count}, we summarize data augmentation strategies of 40 trackers published in recent five years\footnote{Details are listed in the supplementary material. \label{fn:detail}}, and find that most trackers apply random cropping along with several similar combinations of image transformations. Especially for recent transformer-based trackers, all of them follow a similar augmentation pattern as prior CNN-based works. 

Although Bhat~\etal~\cite{updt} has shown the importance of these GDAs on deep CNN models, the efficacy of GDAs has as of yet not been investigated for modern transformer trackers. Hence, we pose the following question: \textbf{What is the impact of GDAs on transformer-based trackers?} To explore the answer, we perform systematic experiments described in Sec.~\ref{sec:experimental_setting}, and analyze results in Sec.~\ref{sec:observation}.
\begin{table}[tbp]
    \centering
    \caption{\textbf{Usage count of each data augmentation in trackers published in five years.} (``RC" indicates random cropping.)}
    \label{tab:count}
    \vspace{-1em}
    \resizebox{0.8\linewidth}{!}{
    \setlength{\tabcolsep}{0.5mm}{
    \begin{tabular}{ccccccc}
    \toprule
    Models & Grayscale &RC&Flip&Bright&Blur&Rotate \\\midrule
        CNN (28)& 10&28&4&23&13&1\\
        Transformer (12)&11&12&8&11&0&0\\
        \bottomrule
    \end{tabular}}}
\vspace{-1.5em}
\end{table}

\begin{figure*}[tbp]
    \centering
    \includegraphics[width=0.99\textwidth]{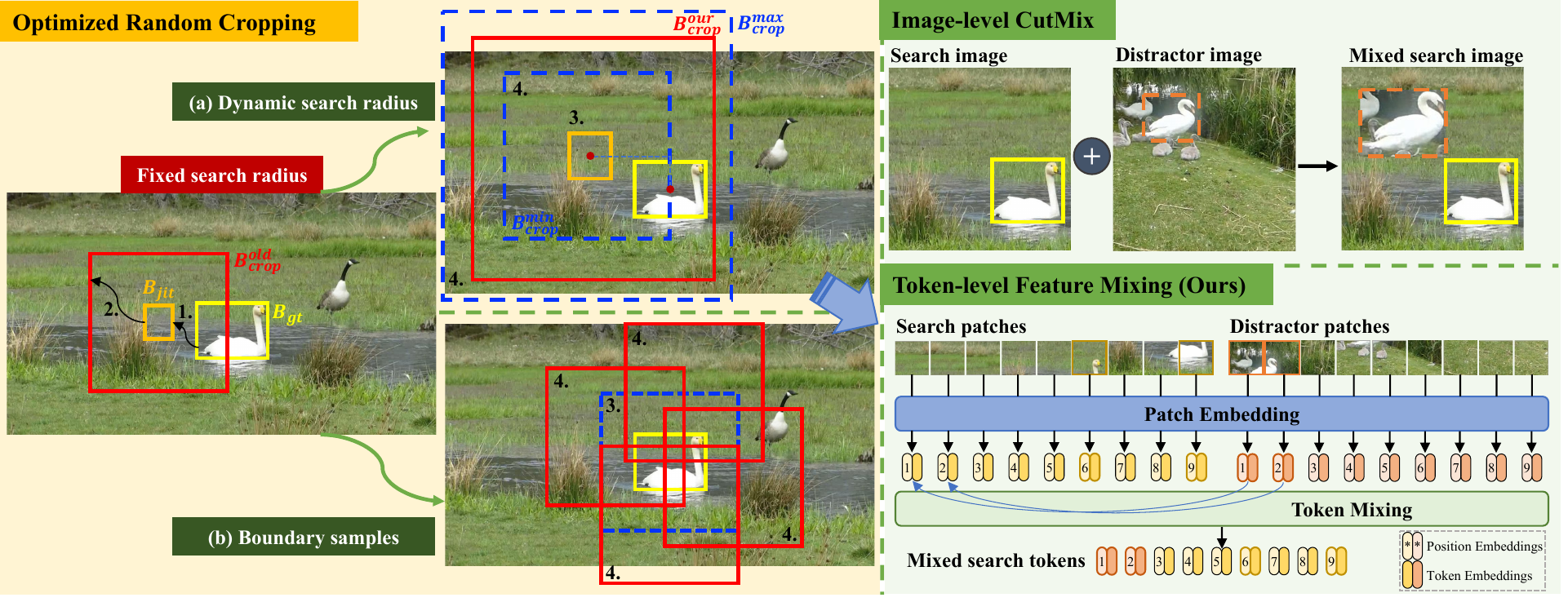}
    \caption{\textbf{Illustration of our customized data augmentation strategies.} Left: Optimized random cropping including (a) dynamic selection mechanism for the search radius factor $\gamma$, and (b) simulation of boundary samples. It does not only enrich the diversity of samples from two-step randomness, but renders also the model insensitive to the parameters. Numbers from \textbf{1.} to \textbf{4.} indicate the order of cropping steps. Right: Classical image-level CutMix (top), and our token-level feature mixing augmentation (bottom).}
    \label{fig:framework}
    \vspace{-1em}
\end{figure*}
\subsection{Experimental Settings}
\label{sec:experimental_setting}
As shown in Fig.~\ref{fig:aug}, we choose OSTrack~\cite{ye2022ostrack} (see (a.1) and (a.2)) and STARK~\cite{yan2021stark} (see (b)) as baselines to represent the pure transformer tracker and the hybrid CNN-Transformer tracker, respectively,
and evaluate all models on LaSOT~\cite{lasot} and LaSOT\_EXT~\cite{fan2021lasotext}. Their official augmentations are grayscale, random cropping, brightness jitter, and horizontal flip, which are also most used for other transformer-based trackers (see Tab.~\ref{tab:count}). Since grayscale is required to make the model robust to grayscale sequences, while random cropping prevents models from overfitting, we consider the two approaches as necessary. The model trained with only these two approaches is represented as ``No", while the official setting is denoted as ``Base". 

To explore the impact of the other methods, along with blur and rotation which are applied by some CNN-based trackers, we remove (horizontal flip or brightness jitter) or add (blur or rotation) each augmentation on the basis of ``Base", represented as ``-Flip", ``-Br", ``+Blur", and ``+Rt" in Fig.~\ref{fig:aug} (a.1) and (b), respectively. Considering stability, we run each experiment three times with different random seeds, and record their average performance with the standard deviation (STD), which is illustrated as error bars.
Besides, to avoid negative effects from inappropriately tuned probability and magnitude, we test different settings\footref{fn:detail} for blur and rotation, and use the best-performing parameters in our systematic experiments.

In addition, we also investigate the impact of the jitter degree of random cropping. In Fig.~\ref{fig:aug} (a.2), we set different jitter degree of random cropping by adjusting the magnitude of scale and shift. The size of circles represents the value of scale\footnote{The scale value is traversed from 0.15 to 0.45}, and the black dotted line indicates the official setting of the baseline (OSTrack).

\subsection{Observations and Analysis}
\label{sec:observation}
\textbf{Different types of GDA.} Experiments in Fig.~\ref{fig:aug} (a.1) and (b) imply that these GDAs seem to have limited effectiveness for the SOTA transformer-based tracker. Taking results on different benchmarks and error bars into account, we can conclude that these GDAs do not provide substantial improvement but only slight fluctuations up and down the baseline models.


\textbf{Different jitter degree of random cropping.} From the trend of dotted curves in Fig.~\ref{fig:aug} (a.2), we find that a proper setting of random cropping can significantly improve the tracking model. The model can benefit more from larger jitter, \eg~$\rm Shift_4$ vs. $\rm Shift_2 $. However, further increasing the jitter degree will cause model degradation, \eg~$\rm Shift_5$.

\textbf{Analysis.} Due to global correlation and models~\cite{he2022mae} pretrained on large-scale datasets, transformer models trained without GDA (see ``No" in Fig.~\ref{fig:aug} (a.1) and (b)) can already address most situations which are difficult for CNNs. However, we observe that challenges like background interference are still difficult for modern trackers (see Fig.~\ref{fig:scoremap}), and cannot be simulated by aforementioned GDAs. 
Therefore, customized augmentations based on unsolved challenges are needed to further improve SOTA transformer trackers.

As for random cropping, we can conclude from the dotted curves in Fig.~\ref{fig:aug} (a.2) that various samples with different target positions and scales are conducive to training models. However, in the existing random cropping strategy with fixed search radius factor $\gamma_{\rm fix}$, the shift parameter should not be set larger than $\gamma_{\rm fix}$ to avoid uninformative samples,~\ie the object is outside the patch. Otherwise, these uninformative samples would pollute the training set and cause model degradation, \eg~results of $\rm Shift_5$ where $\gamma_{\rm fix}=4$. Therefore, the existing random cropping strategy with a fixed context scope, does not only limit the diversity of samples, but also cause the parameter sensitivity.

\section{Data Augmentation Customized for VOT}
\label{sec:datr}
Motivated by the analysis in Sec.~\ref{sec:observation}, we propose two customized data augmentation approaches. First, optimized random cropping (ORC) is proposed, including dynamic selection for search radius factor, and simulation of boundary samples, where the partial region of the object stays at the boundary of the search patch. Second, we propose a token-level feature mixing strategy (TFMix) to simulate unsolved challenges, like background interference. We describe these two augmentations in Sec.~\ref{sec:orc} and Sec.~\ref{sec:cutmix}, respectively.

\subsection{Optimized Random Cropping}
\label{sec:orc}
Existing trackers essentially treat tracking as a local matching problem between templates and search regions. The local search region is decided by the predicted target location of the previous frame, and a fixed search radius factor $\gamma$. To maintain the distribution consistency of samples, random cropping with the same value of $\gamma$ as inference is applied in the training phase. Consider for instance the cropping strategy in the transformer-based methods as an example\footnote{Prior Siamese-based trackers~\cite{SiamFC,siamrpn++,Ocean_2020_ECCV} apply similar parameter as $\gamma$ to fix the context scope.}, as shown in Fig.~\ref{fig:framework} (left), the existing random cropping strategy has two steps, \ie, jitter the groundtruth $B_{\rm gt}$ via random shifting and scaling (``\textbf{1.}" in Fig.~\ref{fig:framework}), and crop the search region $B_{\rm crop}^{\rm old}$ based on the jittered bounding box $B_{\rm jit}$ as well as a fixed $\gamma$ (``\textbf{2.}" in Fig.~\ref{fig:framework}). 

There are several disadvantages to this strategy. First, only one random step (``\textbf{1.}" in Fig.~\ref{fig:framework}) is performed to support diversity of samples. Second, the degree of shift is constraint by $\gamma_{\rm fix}$ to avoid uninformative samples, which leads the training process to be sensitive to parameters. In addition, training with a fixed $\gamma$ makes the model inflexible, \ie~forcing the model to be specific to the same $\gamma$ in inference, shown as Tab.~\ref{tab:sf_inference} (discussed in Sec.~\ref{sec:ablation}).


In this paper, we propose an optimized random cropping strategy to address these issues. 
As shown in Fig.~\ref{fig:framework} (a), to enrich the diversity of training samples, and also unbind the model from the strong association with specific $\gamma$ in inference, we first turn the fixed $\gamma$ during training into a dynamic selected value from $\gamma_{\rm 
 min}$ to $\gamma_{\rm max}$. 
The maximum and minimum values are used to limit the proportion of context in search regions. 
Otherwise, the target will be very small or large in the resized search patch. 
Furthermore, to avoid uninformative samples, we calculate the practical minimum search radius factor $\gamma_{\rm min}^p$ based on the distance between center locations of $B_{\rm gt}$ and $B_{\rm jit}$. 
If $\gamma_{\rm min}^p$ is larger than $\gamma_{\rm max}$, we consider the current $B_{\rm jit}$ to be invalid, and retry to find a proper one. 
Through this simple strategy, uninformative samples can be avoided without scarifying the diversity of the training set.
It is worth noting that although the original random cropping strategy can achieve context variation implicitly by $B_{\rm jit}$, compared with this one-step randomness, our method consists of two random steps, \ie~$B_{\rm jit}$ and dynamic $\gamma$, which are able to obtain samples with more diverse scales and contexts. Qualitative comparisons of sample diversity are presented in the supplementary material.

\begin{algorithm}[!htb]
\caption{Optimized Random Cropping}
\label{alg:Optimized_random_cropping}
\begin{algorithmic}[1]
\Require 
    $I_s$, $B_{\rm gt}$, $\gamma_{\rm min}$, $\gamma_{\rm max}$, $D_{\rm jit}$, $S_{\rm jit}$, $P_{b}$ 
\Ensure
    $B_{\rm crop}$, $\gamma$
\State $\gamma$=Random($\gamma_{\rm min}$,$\gamma_{\rm max}$);
\If{Random(0,1) $< P_{b}$}
        \State $B_{\rm crop}$ = CenterCrop($B_{\rm gt}$, $\gamma$); \Comment{Fig.~\ref{fig:framework} (b) \textbf{``3."}} 
        \State direction = Random(top, bottom, left, right);
        \State $B_{\rm crop}$ = Move($B_{\rm crop}$, direction); \Comment{Fig.~\ref{fig:framework} (b) \textbf{``4."}}
\Else
        \While{True}
            \State $B_{\rm jit}$ = Jitter($B_{\rm gt}$, $D_{\rm jit}$, $S_{\rm jit}$); \Comment{Fig.~\ref{fig:framework} (a) \textbf{``3."}}
            \State $\gamma_{\rm min}^{p}$=MAX\{$\frac{2\vert ct_s - ct_{\rm jit}\vert_{\rm max}}{\sqrt{w_{\rm jit}\times h_{\rm jit}}}$, $\gamma_{\rm min}$\};
            \If{$\gamma_{\rm min}^{p} \leq \gamma_{\rm max}$}
                \State $\gamma$ = Random($\gamma_{\rm min}^{p}$,$\gamma_{\rm max}$); \Comment{Fig.~\ref{fig:framework} (a) \textbf{``4."}}
                \State $B_{\rm crop}$ = CenterCrop($B_{\rm jit}$,$\gamma$);
                \State Break;
            \EndIf
        \EndWhile
\EndIf
\end{algorithmic}
\end{algorithm}
Besides, considering that objects often appear at the boundary or even partially outside search regions in some failure cases and challenges like fast motion. 
we simulate such boundary samples with probability $P_b$, shown as Fig.~\ref{fig:framework} (b). 
We first calculate the search region (blue dashed box) based on $B_{\rm gt}$, and then shift it to a random direction until the target is partially at the boundary. 
It helps models cope with boundary targets more accurately.

The procedure of our ORC is described as Algorithm~\ref{alg:Optimized_random_cropping}. $I_s$ denotes the processed search frame, $D_{\rm jit}$ and $S_{\rm jit}$ represent the magnitude of random shifting and scaling, $P_{b}$ identifies the probability of boundary samples, $ct_s$ and $ct_{\rm jit}$ represent center locations of $B_{\rm gt}$ and $B_{\rm jit}$. 
Due to dynamic $\gamma$ and boundary samples, our ORC can enrich the diversity of samples from different perspectives, such as the context scope, target positions and scales, while avoiding uninformative samples. Experiments in Sec.~\ref{sec:experiments} demonstrate ORC improvements to performance and $\gamma$ flexibility.


\subsection{Token-level Feature Mixing}
\label{sec:cutmix}
Background interference is one of the main challenges for modern trackers, but such samples are not the focus of GDAs, which might be a potential reason for their limited effectiveness. 
Recent augmentations like CutMix~\cite{yun2019cutmix} can be an option to synthesize hard samples with background interference, as shown in Fig.~\ref{fig:framework} (top-right). 
However, such image mixing tends to trap the model in overfitting to sharp border effect. To mitigate this issue and consider the token mechanism of transformer models, we propose a token-level feature mixing method as shown in Fig.~\ref{fig:framework} (bottom-right). 
A search patch with the object $\mathcal{O}_s$, and a distractor patch with another object $\mathcal{O}_d$ are first cropped and processed by a linear projection, we then transfer distractor tokens $\rm T_d^{\mathcal{O}_d}$ belonging to $\mathcal{O}_d$ and replace search tokens $\rm T_s^{\mathcal{O}_d}$ in corresponding positions, represented as 
\begin{equation}
    \centering
    \label{equ:cutmix}
    {\rm T_s^{\mathcal{O}_d}}=\frac{{\rm T_d^{\mathcal{O}_d}}-\mathrm{mean}_{\mathcal{O}_d}}{\mathrm{std}_{\mathcal{O}_d}}\mathrm{std}_{\mathcal{O}_s}+\mathrm{mean}_{\mathcal{O}_s}.
\end{equation}
Distractor tokens $\rm T_d^{\mathcal{O}_d}$ will be normalized before transferring to alleviate huge discrepancy between $\mathcal{O}_s$ and $\mathcal{O}_d$, where $\mathrm{mean}_{\mathcal{O}_{d/s}}$ and $\mathrm{std}_{\mathcal{O}_{d/s}}$ represent the global mean and standard deviation of the object tokens.
To increase the difficulty of samples, we preferentially select $\mathcal{O}_d$ from the same category as $\mathcal{O}_s$. Besides, an occluded threshold is used to control the occluded degree of $\mathcal{O}_s$.

\section{Experiments}
\label{sec:experiments}
To investigate the effectiveness of our augmentations \datr, we apply them to two SOTA transformer trackers, MixFormer~\cite{cui2022mixformer} and OSTrack~\cite{ye2022ostrack}.
Besides, we also apply our ORC to a hybrid CNN-Transformer tracker STARK~\cite{yan2021stark}, and a CNN-based tracker SiamFC++~\cite{xu2020siamFC++} to demonstrate its generalization ability to CNN backbones.

\subsection{Implementation Details}
We implement our data augmentations in Python with PyTorch. All experiments are trained using four NVIDIA A100 GPUs. For our data augmentations, we set the probability of boundary samples, $P_b$, to 0.05. To keep the dynamic selection range of the search radius factor $\gamma$ symmetrical to the fixed value in inference, we set it as [2,6] when $\gamma = 4$ in inference, and [4,6] when $\gamma = 5$. As for the mixing, our TFMix augmentation is triggered every 11 epoches, and the occluded threshold is set to 0.5.
It is worth noting that our \datr~can augment both video and image datasets. For a fair comparison, we adopt the same training settings as for the baseline to retrain each tracking model with and without our augmentations, where training datasets include LaSOT, GOT-10k~\cite{huang2019got10k}, TrackingNet~\cite{trackingnet}, and COCO~\cite{lin2014COCO}.


\subsection{Ablation Study and Analysis}
\label{sec:ablation}
Using OSTrack with 256 image resolution as the baseline tracker, we perform a series of ablation study on LaSOT and LaSOT\_EXT to demonstrate the effectiveness of our approaches from different aspects. LaSOT contains 280 sequences with 70 categories, which are the same as its training subset. In contrast, LaSOT\_EXT is composed of 150 very challenging sequences with 15 unseen categories. 
One-pass evaluation is performed on both benchmarks with three metrics: the success rate (AUC), precision (Pre), and normalized precision ($\rm P_{norm}$). AUC represents the ratio of successfully tracked frames, while Pre and $\rm P_{norm}$ represent the distance of center points between the groundtruth and the predicted result. $\rm P_{norm}$ is normalized with the target scale, which is stable to target size and image resolution.

\textbf{Impact of each component.} As shown in Tab.~\ref{tab:component}, our optimized random cropping can obtain 2.0\% and 0.9\% AUC gains (Base vs. \ding{173}) on LaSOT\_EXT and LaSOT, respectively. Among them, dynamic $\gamma$ mechanism increases AUC with 1.7\% on LaSOT\_EXT, and simulating boundary samples can further improve AUC to 69.3\% on LaSOT. Due to generating challenging samples, our TFMix boosts the performance to 49.7\% on LaSOT\_EXT. 
\begin{table}[tbp]
\centering
\caption{\textbf{Impact of each proposed component on AUC\protect\footnotemark.}}
\label{tab:component}
\vspace{-1em}
\resizebox{0.9\linewidth}{!}{
\setlength{\tabcolsep}{0.8mm}{
\begin{tabular}{cccc|c|c}
\toprule
Method & Dynamic $\gamma$ & Boundary & TFMix & LaSOT & LaSOT\_EXT \\\midrule
Base & &        &        & 68.4  & 47.0  \\
\ding{172}&\ding{52}    & &    & 68.9&  48.7  \\
\ding{173}& \ding{52} & \ding{52}  &    & 69.3 & 49.0  \\
\ding{174}& \ding{52} &\ding{52} &\ding{52} & 69.3 &49.7\\\bottomrule
\end{tabular}}}
\end{table}
\begin{table}[tbp]
\centering
\caption{\textbf{Comparison of different mixing strategies\protect\footref{fn:seed},} including different image-level mixing methods, and late feature mixing.}
\label{tab:cutmix}
\vspace{-1em}
\resizebox{0.9\linewidth}{!}{
\setlength{\tabcolsep}{1.0mm}{
\begin{tabular}{ccccc|ccc}
\toprule
\multicolumn{2}{c}{\multirow{2}{*}{Mixing Strategy}} & \multicolumn{3}{c|}{LaSOT}                   & \multicolumn{3}{c}{LaSOT\_EXT} \\ 
            &        & AUC  & $\rm P_{norm}$ & Pre & AUC   & $\rm P_{norm}$  & Pre     \\ \midrule
\multirow{3}{*}{Image}&Bbox &69.0 &79.0 &75.2  &48.6  &59.3 &55.5 \\
&Mask &69.4 & 79.7&75.8  &48.3 &58.9  & 55.1 \\ 
&Token &69.0   &79.1 & 75.2&49.0&59.7  &55.7   \\\midrule
\multirow{2}{*}{Feature}&Late & 68.8  &79.0 &75.1 &49.0& 59.7 &55.4 \\ 
&Early (Ours) & 69.3  & 79.3&75.3 &49.7& 60.4 &56.6\\\bottomrule
\end{tabular}}}
\end{table}
\footnotetext{Reported results are averaged over 3 random seeds.\label{fn:seed}}

\begin{figure}[tbp]
    \centering
    \vspace{-0.5em}
    \includegraphics[width=0.95\linewidth]{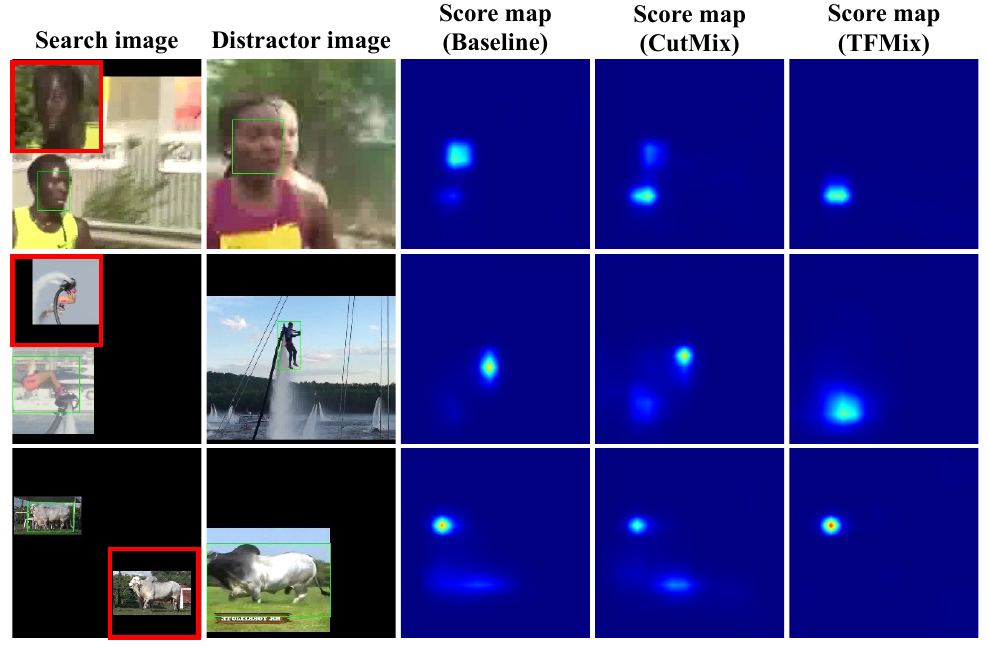}
    
    \caption{\textbf{Comparison of discriminative ability between image-level CutMix and TFMix.} Templates are framed by red boxes.}
    \label{fig:mask_token}
    \vspace{-1.5em}
\end{figure}

\textbf{Different mixing strategies.} 
To demonstrate the effectiveness of the proposed TFMix, we compare it with different mixing strategies, including several different image-level mixing methods, and late feature-level mixing.
The same parameter settings are used for a fair comparison.
For the image mixing with bbox, we simply use bounding box annotations to crop a rectangle area of the distractor, while for the image mixing with mask, we first obtain mask annotations of all training datasets from Alpha-Refine~\cite{yan2021alpha}, and paste the distractor itself without any context. The token image mixing is similar to TokenMix~\cite{liu2022tokenmix}, we set the same token size as for the model, and randomly mix 30\% to 50\% of them between search patches and distractor patches. 
As for the late feature mixing, our TFMix can be considered as an early-stage feature mixing, since the mixing is performed before the feature extraction and fusion. 
In contrast, the late mixing delays this operation until after the feature fusing. In the feature extraction and fusion stage, token interactions not only happen between the template and search patch, but also occur between context tokens in the search patch itself. Therefore, late feature mixing will miss the core interaction between the original object and the extra distractor. 
As shown in Tab.~\ref{tab:cutmix}, our TFMix is superior to other mixing strategies, especially on the most challenging benchmark, LaSOT\_EXT.

Moreover, we compare the discriminative ability gained from image mixing with bbox (CutMix), and our TFMix. As shown in Fig.~\ref{fig:mask_token}, when distractor tokens are mixed into search patches via Eq.~\ref{equ:cutmix}, the baseline tracker is prone to being confused by distractors. CutMix improves this phenomenon to some extent (see first row), while the last column shows that our TFMix promotes the model be more discriminative to distractors.

\begin{table}[tbp]
\centering
\caption{\textbf{Adaptability comparison to different $\gamma$ in inference.}}
\label{tab:sf_inference}
\vspace{-1em}
\resizebox{.9\linewidth}{!}{
\setlength{\tabcolsep}{0.9mm}{
\begin{tabular}{c cccc|cccc}
\toprule
    \multirow{2}{*}{$\gamma_{\text{train}}\to\gamma_{\text{test}}$}  &\multicolumn{4}{c|}{LaSOT}  &   \multicolumn{4}{c}{LaSOT\_EXT} \\ 
          & AUC  & $\rm P_{norm}$ & Pre &STD& AUC   & $\rm P_{norm}$  & Pre &STD    \\ \midrule
\multicolumn{1}{c}{$\rm 4\to3$}  & 58.9&67.5 &64.9 &\multirow{3}{*}{\textbf{5.02}} &40.1 &49.4&46.1 &\multirow{3}{*}{\textbf{5.12}}   \\
\multicolumn{1}{c}{$\rm 4\to4$}  & 68.6 & 78.1 & 74.4 && 47.3    & 57.4    & 53.1&  \\
\multicolumn{1}{c}{$\rm 4\to5$}  & 61.5&68.7 &63.8 & & 37.4& 44.2 &38.7 & \\ \midrule 
\multicolumn{1}{c}{$\rm 3\to3$}  & 67.5& 76.3&72.9 &\multirow{3}{*}{\textbf{0.64}} &43.6 &52.2& 47.8&\multirow{3}{*}{\textbf{2.40}}   \\
\multicolumn{1}{c}{$\rm 4\to4$}  & 68.6 & 78.1 & 74.4& & 47.3    & 57.4    & 53.1&  \\
\multicolumn{1}{c}{$\rm 5\to5$}  & 67.5& 77.4& 72.7 && 48.1& 58.5 &54.4&  \\ \midrule
\multicolumn{1}{c}{Dyn.$\rm \to 3$}   &67.1 &76.7 &72.6 &\multirow{3}{*}{\textbf{1.11}} &44.2 &54.0  &50.0 &\multirow{3}{*}{\textbf{2.52}}  \\
\multicolumn{1}{c}{Dyn.$\rm \to 4$}   &69.3 & 79.4 & 75.5 && 48.8  &59.4 & 55.6& \\
\multicolumn{1}{c}{Dyn.$\rm \to 5$}   &68.4 &78.3 &73.7 & &48.3 & 58.6 &54.4 &  \\
\bottomrule
\end{tabular}
}}
\vspace{-1.5em}
\end{table}

\textbf{Adaptability to different $\gamma$ in inference.} Different from prior training, since the proposed dynamic $\gamma$ mechanism enriches training samples from the perspective of contextual information, the model should be more adaptive to search patches cropped with different $\gamma$ in inference. To investigate the validity of this conjecture, we conduct three sets of experiments shown as Tab.~\ref{tab:sf_inference}, where ``$\gamma_{\text{train}}\to\gamma_{\text{test}}$" represents that the search radius factor is set to $\gamma_{\text{train}}$ in the training phase, and $\gamma_{\text{test}}$ in the inference. ``Dyn." represents to train the model using our dynamic $\gamma$ mechanism. 

We can see that the model trained with a fixed $\gamma_{\text{train}}$ performs extremely poorly when faced with different $\gamma_{\text{test}}$ in the inference (see results of ``$4\to i$"). The AUC standard deviation (STD) of the first set is higher than 5 on both two benchmarks. While in the second set (``$i\to i$"), well performance with lower STD under different $\gamma_{\text{test}}$ can be obtained when we keep the consistency of $\gamma$ in the training and inference. This phenomenon shows that the original cropping strategy using fixed $\gamma_{\text{train}}$ establishes a strong association of $\gamma$ between the training and inference, which hinders the adaptability of models to scale variations, especially caused by previous inaccurate prediction (see Fig.~\ref{fig:scoremap} (left)).

In contrast, our model effectively unbinds this kind of association due to the proposed dynamic $\gamma$ in the training phase. Our model (``Dyn. $\to i$") performs well on all different $\gamma_{\text{test}}$, and has a comparable low STD with the second set. We think this characteristic of our approach not only helps the model to be more robust to scale variations, but also provides a new insight for future works related to dynamic search in the inference, like~\cite{zhu2022srrt}.

\textbf{Stability to different magnitudes of jitter.} As concluded in Sec.~\ref{sec:observation}, tracking models cannot perform well under small and very large jitter settings in the training phase. To demonstrate that our ORC is more stable to different jitter degrees, we train our model under different jitter settings, as shown in Fig.~\ref{fig:aug} (a.2). Compared with the original cropping method (light dashed lines), our ORC (dark solid lines) enables the tracking model adapt to varying degrees of jitter. In addition to dynamic $\gamma$ mechanism, which enriches samples' diversity, simulating boundary cases can feed models such samples under a small jitter setting. Besides, there is also a check and filter step for uninformative samples in our ORC. Therefore, we can still obtain well-performed model stably even under very small (\eg$\rm Shift_2$) or very large (\eg$\rm Shift_5$) jitter. We think this characteristic brings convenience for future works, which prevents models from being too sensitive to jitter parameters.

\begin{table}[tbp]
    \centering
    \caption{\textbf{Generalization of our methods to CNN backbones.}}
    \vspace{-1em}
    \label{tab:general1}
    \resizebox{.8\linewidth}{!}{
\setlength{\tabcolsep}{0.5mm}{
    \begin{tabular}{c|ccc|cc}
    \toprule
        AUC & STARK&+ORC&+TFMix&SiamFC++&+ORC \\ \midrule
        LaSOT & 66.4&67.7&66.2&60.4&61.1\\
        LaSOT\_EXT&46.5&48.1&46.9&37.7&38.9\\ \bottomrule
    \end{tabular}}}
    \vspace{-0.5em}
\end{table}

\textbf{Generalization capability of our methods.} As shown in Tab.~\ref{tab:general1}, in addition to the pure Transformer trackers, our ORC also boosts hybrid CNN-Transformer trackers (\eg STARK) and CNN-based trackers (\eg SiamFC++). However, since our TFMix relies on characteristics of transformer models, \ie global correlation between independent tokens, it shows to be less effective for CNN backbones, causing an average 1.4\% AUC decline for STARK. The potential reason might be the strong inductive bias in CNN networks. Detailed explanations and experimental settings can be found in the supplementary material.


\begin{table}[tbp]
\centering
\caption{\textbf{Performance comparison on VOT2022 benchmark.}}
\label{tab:vot}
\vspace{-1em}
\resizebox{0.8\linewidth}{!}{%
\begin{tabular}{clcc}
\toprule
               & EAO  & A  &  R \\ \midrule
\rowcolor{blue!5}MixFormer-22k  & 0.538& 0.776 & 0.838 \\
\rowcolor{blue!12}+\textbf{\datr} (Ours)&0.531~\tiny{\color{gray}$0.7\%\downarrow$} &  0.743&0.840 \\
\rowcolor{orange!10}OSTrack256&0.497 & 0.783 & 0.788\\
\rowcolor{orange!20}+\textbf{\datr} (Ours)&0.525~\tiny{\color{gray}$2.8\%\uparrow$} & 0.771& 0.820\\
\rowcolor{blue!5}OSTrack384 &0.522 & 0.788 &0.799 \\
\rowcolor{blue!12}+\textbf{\datr} (Ours) &0.525~\tiny{\color{gray}$0.3\%\uparrow$} & 0.777 & 0.807\\
\midrule
Average gain&+0.8\%&-1.9\%&+1.4\%\\
 \bottomrule
\end{tabular}}
\vspace{-1.3em}
\end{table}

\begin{table*}[!htb]
\centering
\caption{\textbf{State-of-the-art comparisons on five tracking benchmarks.} The top two results are highlighted with \textbf{\color{red}red} and \textbf{\color{blue}blue}, respectively.}
\vspace{-0.5em}
\label{tab:sota}
\resizebox{0.95\linewidth}{!}{%
\setlength{\tabcolsep}{1.0mm}{
\begin{tabular}{c|llc|llc|llc|ll|ll}
\toprule
\multirow{2}{*}{Method}  & \multicolumn{3}{c|}{LaSOT} & \multicolumn{3}{c|}{LaSOT\_EXT} & \multicolumn{3}{c|}{GOT-10k} & \multicolumn{2}{c|}{UAV123} & \multicolumn{2}{c}{NFS} \\
                 &    AUC  & $\rm P_{norm}$ & Pre  & AUC     & $\rm P_{norm}$    & Pre     & AO     & $\rm SR_{0.5}$    & $\rm SR_{0.75}$ & AUC     & Pre & AUC     & Pre   \\ \midrule
                 
ECO~\cite{eco}& 32.4 & 33.8     & 30.1 & 22.0    & 25.2        & 24.0    & 31.6    & 30.9        & 11.1  &53.5&76.9&52.2&63.4  \\
SiamFC~\cite{SiamFC}& 33.6 & 42.0     & 33.9 & 23.0    & 31.1        & 26.9    & 34.8    & 35.3        & 9.8 &46.8&69.4&37.7&44.5   \\
MDNet~\cite{MDNet} & 39.7 & 46.0     & 37.3 & 27.9    & 34.9        & 31.8    & 29.9    & 30.3        & 9.9  &52.8&-&42.2&-  \\
SiamRPN++~\cite{siamrpn++}& 49.6 & 56.9     & 49.1 & 34.0    & 41.6        & 39.6    & 51.7    & 61.6        & 32.5  &59.3&78.2&57.1&69.3  \\
Ocean~\cite{Ocean_2020_ECCV}& 56.0 & 65.1     & 56.6 & -       & -           & -       & 61.1      & 72.1      & 47.3  &57.4&77.8&49.4&61.2     \\
DiMP~\cite{dimp}& 56.9 & 65.0     & 56.7 & 39.2    & 47.6        & 45.1    & 61.1    & 71.7        & 49.2  &64.3&85.1&61.8&73.8  \\
TrDiMP~\cite{wang2021transformer}&63.9 & -        & 61.4 & -       & -           & -       & 67.1  & 77.7        & 58.3  &66.4&86.9&66.2&79.1     \\
SiamRCNN~\cite{voigtlaender2020siamrcnn}&64.8  & 72.2 &-&-& - & - & 64.9 &- &- &64.9&83.4&63.9&- \\
TransT~\cite{chen2021transt}& 64.9 & 73.8     & 69.0 & -       & -           & -       & 67.1     & 76.8   & 60.9 &68.1&87.6&65.3&78.8   \\
SBT-L~\cite{xie2022SBT}& 66.7 & - & 71.1 & -        & -            & -        & 70.4 & 80.8 & 64.7 &-&-&-&- \\
KeepTrack~\cite{mayer2021keeptrack}& 67.1 & 77.2     & 70.2 & 48.2    & -           & -       & -       & -           & -  &69.7&-&66.4& -    \\
ToMP-101~\cite{mayer2022transforming}& 68.5& 79.2 & 73.5 & 45.9 & - &- & - & -& - &66.9&-&\secondresult{66.7}&-  \\
AiATrack~\cite{gao2022aiatrack}& 69.0 & 79.4 &73.8 &46.8  & 54.4& 54.2 &69.6 &  80.0 & 63.2&70.6&-&\topresult{67.9}&-  \\
Sim-L~\cite{chen2022simtrack}& 70.5 & 79.7 &- &-  & -& - &69.8 &  78.8 & 66.0&\topresult{71.2}&\secondresult{91.6}&-&-  \\ \midrule
\rowcolor{blue!5}MixFormer-22k~\cite{cui2022mixformer}&68.9 & 78.5 &74.3     & 49.1 & 59.6 & 55.3 &70.3 & 80.0 & 66.2 &69.7&91.0&65.0&79.1    \\
\rowcolor{blue!12}+\textbf{\datr} (Ours) & \bestresult{68.8}~\tiny{\color{gray}$0.1\downarrow$}&\bestresult{78.9}~\tiny{\color{gray}$0.4\uparrow$} & \bestresult{74.6} & \secondresult{51.0}~\tiny{\color{gray}$1.9\uparrow$} & \secondresult{61.8}~\tiny{\color{gray}$2.2\uparrow$} & \bestresult{57.3} & \bestresult{71.4}~\tiny{\color{gray}$1.1\uparrow$} & 81.0~\tiny{\color{gray}$1.0\uparrow$} & \bestresult{67.6} &\bestresult{69.6}~\tiny{\color{gray}$0.1\downarrow$}&90.9~\tiny{\color{gray}$0.1\downarrow$}&\bestresult{65.8}~\tiny{\color{gray}$0.8\uparrow$}&\bestresult{79.7}~\tiny{\color{gray}$0.6\uparrow$}   \\
\rowcolor{orange!10}OSTrack256~\cite{ye2022ostrack}& 68.6 & 78.1     & 74.4 & 47.3    & 57.4        & 53.1    &71.4 & 81.4  & 67.5 &68.2&88.6&65.4&79.6    \\
\rowcolor{orange!20}+\textbf{\datr} (Ours) &69.1~\tiny{\color{gray}$0.5\uparrow$} & 79.1~\tiny{\color{gray}$1.0\uparrow$} & 75.2 &  49.9~\tiny{\color{gray}$2.6\uparrow$}  & 60.6~\tiny{\color{gray}$3.2\uparrow$}   &  57.0 & 72.5~\tiny{\color{gray}$1.1\uparrow$}  &  82.3~\tiny{\color{gray}$0.9\uparrow$}&69.2&\secondresult{70.8}~\tiny{\color{gray}$2.6\uparrow$}&\topresult{92.4}~\tiny{\color{gray}$3.8\uparrow$}&66.0~\tiny{\color{gray}$0.6\uparrow$}&\topresult{81.1}~\tiny{\color{gray}$1.5\uparrow$}    \\
\rowcolor{blue!5}OSTrack384~\cite{ye2022ostrack}& \secondresult{70.7} & \secondresult{80.4}     & \secondresult{77.0} & 50.5    & 61.2        & \secondresult{57.4}    &\secondresult{73.5} & \secondresult{83.0}  & \secondresult{70.6} &69.7&90.6&66.3&\secondresult{80.8}    \\
\rowcolor{blue!12}+\textbf{\datr} (Ours) &\topresult{71.0}~\tiny{\color{gray}$0.3\uparrow$} & \topresult{80.7}~\tiny{\color{gray}$0.3\uparrow$} & \topresult{77.5} &  \topresult{51.8}~\tiny{\color{gray}$1.3\uparrow$}  & \topresult{62.7}~\tiny{\color{gray}$1.5\uparrow$}   &  \topresult{59.0} & \topresult{74.2}~\tiny{\color{gray}$0.7\uparrow$}  &  \topresult{84.1}~\tiny{\color{gray}$1.1\uparrow$}&\topresult{71.1}&69.7~\tiny{\color{gray}$0.0\uparrow$}&\bestresult{90.7}~\tiny{\color{gray}$0.1\uparrow$}&65.5~\tiny{\color{gray}$0.8\downarrow$}&79.9~\tiny{\color{gray}$0.9\downarrow$}    \\
\midrule
Average gain&+0.2\% & +0.6\% & +0.5\% & +1.9\% & +2.3\% & +2.6\% & +1.0\% & +1.0\% & +1.2\% & +0.8\% & +1.3\% & +0.2\% & +0.4\%\\
 \bottomrule
\end{tabular}
}}
\vspace{-1em}
\end{table*}

\subsection{State-of-the-art comparison}
We apply our augmentations on two SOTA transformer trackers, MixFormer and OSTrack, and evaluate them on six tracking benchmarks. For the OSTrack, we evaluate its two variants with different image resolutions, represented as OSTrack256 and OSTrack384, respectively.

\textbf{VOT2022 (STB)~\cite{vot2022}.} This challenge contains 50 challenging short-term sequences with multiple initial anchor points. The primary measure is the expected average overlap (EAO), which is a principled combination of tracking accuracy (A) and robustness (R). As shown in Tab.~\ref{tab:vot}, our \datr~improves three baseline models by 0.8\% on average in terms of EAO, especially for OSTrack256, boosting by 2.8\% EAO. We can see that our \datr~mainly improves models from the perspective of tracking robustness.

\textbf{LaSOT and LaSOT\_EXT.} Compared with LaSOT, its extended dataset LaSOT\_EXT is more challenging, and all its categories are unseen from the training set. As shown in Tab.~\ref{tab:sota}, the superiority of our augmentations can be fully reflected on the very challenging benchmark LaSOT\_EXT. All of three baseline trackers are improved by 1.9\% AUC and 2.3\% $\rm P_{norm}$ on average. Our augmentations also bring an average of 0.6 $\rm P_{norm}$ gain on LaSOT.

\textbf{GOT-10k.} GOT-10k is composed of 180 test sequences of which classes are zero-overlapped with its training set. We follow the official one-shot protocol to train all models, where only its training subset is allowed to be used for training. Performance is evaluated by three metrics: average overlap (AO), and success rates with two different thresholds ($\rm SR_{0.5}$ and $\rm SR_{0.75}$). As shown in Tab.~\ref{tab:sota}, all of our models achieve significant promotion, surpassing baseline trackers by 1.0\% AO and 1.2\% $\rm SR_{0.75}$ on average.

\textbf{UAV123~\cite{uav123} and NFS~\cite{kiani2017nfs}.} These two benchmarks contain 123 sequences captured from the aerial perspective, and 100 sequences, respectively. Results in Tab.~\ref{tab:sota} show that our \datr~obtains 1.3\% improvement in terms of precision on UAV123, and also minor increase on NFS.

\textbf{Discussion.}
In terms of the above experiments, the superiority of our \datr~is most evident under challenging settings, like dealing with unseen classes (GOT-10k) or very challenging sequences (LaSOT\_EXT), and handling images with small resolution (OSTrack256). A more quantitative analysis of the performance bias on different benchmarks and models, additional qualitative results, and attribute analysis are presented in the supplementary material.

\section{Conclusion}
In this paper, we systematically analyze the impact of GDAs on modern transformer trackers and propose two customized data augmentations for VOT. First, to improve the adaptability of models to scale variations and boundary targets, we design an optimized random cropping, containing dynamic selection for search radius factor, and simulation of boundary samples. Second, we synthesize hard samples with background interference by a token-level feature mixing strategy. Extensive experiments on two SOTA transformer-based trackers and six benchmarks demonstrate our augmentations enable the model benefit from more diverse and challenging samples, and be more flexible to changes of search radius in inference.

\noindent \textbf{Limitation.} Since our augmentations are motivated by unsolved challenges and failure cases, our \datr~tends to improve models in terms of tracking robustness, instead of accuracy, \ie~we aim to locate the target successfully under challenging situations. This might also be the potential reason for the slight accuracy decline in Tab.~\ref{tab:vot}, and minor performance gains on some benchmarks, like LaSOT and NFS.

{\small
\bibliographystyle{ieee_fullname}
\bibliography{references}

\begin{thebibliography}{10}\itemsep=-1pt

\bibitem{SiamFC}
Luca Bertinetto, Jack Valmadre, Joao~F Henriques, Andrea Vedaldi, and Philip~HS
  Torr.
\newblock Fully-convolutional siamese networks for object tracking.
\newblock In {\em European Conference on Computer Vision}, pages 850--865,
  2016.

\bibitem{dimp}
Goutam Bhat, Martin Danelljan, Luc~Van Gool, and Radu Timofte.
\newblock Learning discriminative model prediction for tracking.
\newblock In {\em IEEE International Conference on Computer Vision}, pages
  6182--6191, 2019.

\bibitem{updt}
Goutam Bhat, Joakim Johnander, Martin Danelljan, Fahad Shahbaz~Khan, and
  Michael Felsberg.
\newblock Unveiling the power of deep tracking.
\newblock In {\em European Conference on Computer Vision}, pages 483--498,
  2018.

\bibitem{bhojanapalli2021understanding}
Srinadh Bhojanapalli, Ayan Chakrabarti, Daniel Glasner, Daliang Li, Thomas
  Unterthiner, and Andreas Veit.
\newblock Understanding robustness of transformers for image classification.
\newblock In {\em IEEE International Conference on Computer Vision}, pages
  10231--10241, 2021.

\bibitem{bolme2010MOSSE}
David~S Bolme, J~Ross Beveridge, Bruce~A Draper, and Yui~Man Lui.
\newblock Visual object tracking using adaptive correlation filters.
\newblock In {\em IEEE Conference on Computer Vision and Pattern Recognition},
  pages 2544--2550, 2010.

\bibitem{chen2022simtrack}
Boyu Chen, Peixia Li, Lei Bai, Lei Qiao, Qiuhong Shen, Bo Li, Weihao Gan, Wei
  Wu, and Wanli Ouyang.
\newblock Backbone is all your need: a simplified architecture for visual
  object tracking.
\newblock In {\em European Conference on Computer Vision}, pages 375--392,
  2022.

\bibitem{chen2021transt}
Xin Chen, Bin Yan, Jiawen Zhu, Dong Wang, Xiaoyun Yang, and Huchuan Lu.
\newblock Transformer tracking.
\newblock In {\em IEEE Conference on Computer Vision and Pattern Recognition},
  pages 8126--8135, 2021.

\bibitem{cubuk2018autoaugment}
Ekin~D. Cubuk, Barret Zoph, Dandelion Mane, Vijay Vasudevan, and Quoc~V. Le.
\newblock {AutoAugment}: Learning augmentation strategies from data.
\newblock In {\em Proceedings of the IEEE/CVF Conference on Computer Vision and
  Pattern Recognition (CVPR)}, June 2019.

\bibitem{cui2022mixformer}
Yutao Cui, Cheng Jiang, Limin Wang, and Gangshan Wu.
\newblock {MixFormer}: End-to-end tracking with iterative mixed attention.
\newblock In {\em IEEE Conference on Computer Vision and Pattern Recognition},
  pages 13608--13618, 2022.

\bibitem{ATOM}
Martin Danelljan, Goutam Bhat, Fahad~Shahbaz Khan, and Michael Felsberg.
\newblock {ATOM}: {A}ccurate tracking by overlap maximization.
\newblock In {\em IEEE Conference on Computer Vision and Pattern Recognition},
  pages 4660--4669, 2019.

\bibitem{eco}
Martin Danelljan, Goutam Bhat, Fahad Shahbaz~Khan, and Michael Felsberg.
\newblock {ECO}: {E}fficient convolution operators for tracking.
\newblock In {\em IEEE Conference on Computer Vision and Pattern Recognition},
  pages 6638--6646, 2017.

\bibitem{dwibedi2017cut}
Debidatta Dwibedi, Ishan Misra, and Martial Hebert.
\newblock Cut, paste and learn: Surprisingly easy synthesis for instance
  detection.
\newblock In {\em IEEE International Conference on Computer Vision}, pages
  1301--1310, 2017.

\bibitem{fan2021lasotext}
Heng Fan, Hexin Bai, Liting Lin, Fan Yang, Peng Chu, Ge Deng, Sijia Yu,
  Mingzhen Huang, Juehuan Liu, Yong Xu, et~al.
\newblock {LaSOT}: A high-quality large-scale single object tracking benchmark.
\newblock {\em International Journal of Computer Vision}, 129:439--461, 2021.

\bibitem{lasot}
Heng Fan, Liting Lin, Fan Yang, Peng Chu, Ge Deng, Sijia Yu, Hexin Bai, Yong
  Xu, Chunyuan Liao, and Haibin Ling.
\newblock {LaSOT}: {A} high-quality benchmark for large-scale single object
  tracking.
\newblock In {\em IEEE Conference on Computer Vision and Pattern Recognition},
  pages 5374--5383, 2019.

\bibitem{gao2022aiatrack}
Shenyuan Gao, Chunluan Zhou, Chao Ma, Xinggang Wang, and Junsong Yuan.
\newblock {AiATrack}: Attention in attention for transformer visual tracking.
\newblock In {\em European Conference on Computer Vision}, pages 146--164,
  2022.

\bibitem{ghiasi2021copypaste}
Golnaz Ghiasi, Yin Cui, Aravind Srinivas, Rui Qian, Tsung-Yi Lin, Ekin~D.
  Cubuk, Quoc~V. Le, and Barret Zoph.
\newblock Simple copy-paste is a strong data augmentation method for instance
  segmentation.
\newblock In {\em IEEE Conference on Computer Vision and Pattern Recognition},
  pages 2918--2928, 2021.

\bibitem{he2022mae}
Kaiming He, Xinlei Chen, Saining Xie, Yanghao Li, Piotr Doll{\'a}r, and Ross
  Girshick.
\newblock Masked autoencoders are scalable vision learners.
\newblock In {\em IEEE Conference on Computer Vision and Pattern Recognition},
  pages 16000--16009, 2022.

\bibitem{kcf}
Jo{\~a}o~F Henriques, Rui Caseiro, Pedro Martins, and Jorge Batista.
\newblock High-speed tracking with kernelized correlation filters.
\newblock {\em IEEE Transactions on Pattern Analysis and Machine Intelligence},
  37(3):583--596, 2014.

\bibitem{huang2019got10k}
Lianghua Huang, Xin Zhao, and Kaiqi Huang.
\newblock {GOT-10k}: A large high-diversity benchmark for generic object
  tracking in the wild.
\newblock {\em IEEE Transactions on Pattern Analysis and Machine Intelligence},
  43(5):1562--1577, 2019.

\bibitem{kalal2011TLD}
Zdenek Kalal, Krystian Mikolajczyk, and Jiri Matas.
\newblock Tracking-learning-detection.
\newblock {\em IEEE Transactions on Pattern Analysis and Machine Intelligence},
  34(7):1409--1422, 2011.

\bibitem{kiani2017nfs}
Hamed Kiani~Galoogahi, Ashton Fagg, Chen Huang, Deva Ramanan, and Simon Lucey.
\newblock Need for speed: A benchmark for higher frame rate object tracking.
\newblock In {\em IEEE International Conference on Computer Vision}, pages
  1125--1134, 2017.

\bibitem{vot2022}
Matej Kristan, Ale{\v{s}} Leonardis, Ji{\v{r}}{\'\i} Matas, Michael Felsberg,
  Roman Pflugfelder, Joni-Kristian K{\"a}m{\"a}r{\"a}inen, Hyung~Jin Chang,
  Martin Danelljan, Luka~{\v{C}}ehovin Zajc, Alan Luke{\v{z}}i{\v{c}}, et~al.
\newblock The tenth visual object tracking vot2022 challenge results.
\newblock In {\em European Conference on Computer Vision}, pages 431--460.
  Springer, 2022.

\bibitem{siamrpn++}
Bo Li, Wei Wu, Qiang Wang, Fangyi Zhang, Junliang Xing, and Junjie Yan.
\newblock {SiamRPN++}: {E}volution of siamese visual tracking with very deep
  networks.
\newblock In {\em IEEE Conference on Computer Vision and Pattern Recognition},
  pages 4282--4291, 2019.

\bibitem{li2023self}
Xin Li, Wenjie Pei, Yaowei Wang, Zhenyu He, Huchuan Lu, and Ming-Hsuan Yang.
\newblock Self-supervised tracking via target-aware data synthesis.
\newblock {\em IEEE Transactions on Neural Networks and Learning Systems},
  2023.

\bibitem{lin2014COCO}
Tsung-Yi Lin, Michael Maire, Serge Belongie, James Hays, Pietro Perona, Deva
  Ramanan, Piotr Doll{\'a}r, and C~Lawrence Zitnick.
\newblock Microsoft coco: Common objects in context.
\newblock In {\em European Conference on Computer Vision}, pages 740--755,
  2014.

\bibitem{liu2022tokenmix}
Jihao Liu, Boxiao Liu, Hang Zhou, Hongsheng Li, and Yu Liu.
\newblock {TokenMix}: Rethinking image mixing for data augmentation in vision
  transformers.
\newblock In {\em European Conference on Computer Vision}, pages 455--471,
  2022.

\bibitem{liu2021swin}
Ze Liu, Yutong Lin, Yue Cao, Han Hu, Yixuan Wei, Zheng Zhang, Stephen Lin, and
  Baining Guo.
\newblock {Swin Transformer}: Hierarchical vision transformer using shifted
  windows.
\newblock In {\em IEEE International Conference on Computer Vision}, pages
  10012--10022, 2021.

\bibitem{mayer2022transforming}
Christoph Mayer, Martin Danelljan, Goutam Bhat, Matthieu Paul, Danda~Pani
  Paudel, Fisher Yu, and Luc Van~Gool.
\newblock Transforming model prediction for tracking.
\newblock In {\em IEEE Conference on Computer Vision and Pattern Recognition},
  pages 8731--8740, 2022.

\bibitem{mayer2021keeptrack}
Christoph Mayer, Martin Danelljan, Danda~Pani Paudel, and Luc Van~Gool.
\newblock Learning target candidate association to keep track of what not to
  track.
\newblock In {\em IEEE International Conference on Computer Vision}, pages
  13444--13454, 2021.

\bibitem{uav123}
Matthias Mueller, Neil Smith, and Bernard Ghanem.
\newblock A benchmark and simulator for {UAV} tracking.
\newblock In {\em European Conference on Computer Vision}, pages 445--461,
  2016.

\bibitem{trackingnet}
Matthias Muller, Adel Bibi, Silvio Giancola, Salman Alsubaihi, and Bernard
  Ghanem.
\newblock {TrackingNet}: {A} large-scale dataset and benchmark for object
  tracking in the wild.
\newblock In {\em European Conference on Computer Vision}, pages 300--317,
  2018.

\bibitem{MDNet}
Hyeonseob Nam and Bohyung Han.
\newblock Learning multi-domain convolutional neural networks for visual
  tracking.
\newblock In {\em IEEE Conference on Computer Vision and Pattern Recognition},
  pages 4293--4302, 2016.

\bibitem{naseer2021intriguing}
Muhammad~Muzammal Naseer, Kanchana Ranasinghe, Salman~H Khan, Munawar Hayat,
  Fahad Shahbaz~Khan, and Ming-Hsuan Yang.
\newblock Intriguing properties of vision transformers.
\newblock {\em Advances in Neural Information Processing Systems},
  34:23296--23308, 2021.

\bibitem{paul2022vision}
Sayak Paul and Pin-Yu Chen.
\newblock Vision transformers are robust learners.
\newblock In {\em AAAI Conference on Artificial Intelligence}, volume~36, pages
  2071--2081, 2022.

\bibitem{voigtlaender2020siamrcnn}
Paul Voigtlaender, Jonathon Luiten, Philip~HS Torr, and Bastian Leibe.
\newblock {Siam R-CNN}: Visual tracking by re-detection.
\newblock In {\em IEEE Conference on Computer Vision and Pattern Recognition},
  pages 6578--6588, 2020.

\bibitem{wang2021transformer}
Ning Wang, Wengang Zhou, Jie Wang, and Houqiang Li.
\newblock Transformer meets tracker: Exploiting temporal context for robust
  visual tracking.
\newblock In {\em IEEE Conference on Computer Vision and Pattern Recognition},
  pages 1571--1580, 2021.

\bibitem{wang2021pyramid}
Wenhai Wang, Enze Xie, Xiang Li, Deng-Ping Fan, Kaitao Song, Ding Liang, Tong
  Lu, Ping Luo, and Ling Shao.
\newblock Pyramid vision transformer: A versatile backbone for dense prediction
  without convolutions.
\newblock In {\em IEEE International Conference on Computer Vision}, pages
  568--578, 2021.

\bibitem{wang2020cnn}
Yong Wang, Xian Wei, Xuan Tang, Hao Shen, and Lu Ding.
\newblock Cnn tracking based on data augmentation.
\newblock {\em Knowledge-Based Systems}, 194:105594, 2020.

\bibitem{xie2022SBT}
Fei Xie, Chunyu Wang, Guangting Wang, Yue Cao, Wankou Yang, and Wenjun Zeng.
\newblock Correlation-aware deep tracking.
\newblock In {\em IEEE Conference on Computer Vision and Pattern Recognition},
  pages 8751--8760, 2022.

\bibitem{xu2020siamFC++}
Yinda Xu, Zeyu Wang, Zuoxin Li, Ye Yuan, and Gang Yu.
\newblock {SiamFC++}: Towards robust and accurate visual tracking with target
  estimation guidelines.
\newblock In {\em AAAI Conference on Artificial Intelligence}, volume~34, pages
  12549--12556, 2020.

\bibitem{yan2021stark}
Bin Yan, Houwen Peng, Jianlong Fu, Dong Wang, and Huchuan Lu.
\newblock Learning spatio-temporal transformer for visual tracking.
\newblock In {\em IEEE International Conference on Computer Vision}, pages
  10448--10457, 2021.

\bibitem{yan2021alpha}
Bin Yan, Xinyu Zhang, Dong Wang, Huchuan Lu, and Xiaoyun Yang.
\newblock {Alpha-Refine}: Boosting tracking performance by precise bounding box
  estimation.
\newblock In {\em IEEE Conference on Computer Vision and Pattern Recognition},
  pages 5289--5298, 2021.

\bibitem{ye2022ostrack}
Botao Ye, Hong Chang, Bingpeng Ma, Shiguang Shan, and Xilin Chen.
\newblock Joint feature learning and relation modeling for tracking: A
  one-stream framework.
\newblock In {\em European Conference on Computer Vision}, pages 341--357,
  2022.

\bibitem{yun2019cutmix}
Sangdoo Yun, Dongyoon Han, Seong~Joon Oh, Sanghyuk Chun, Junsuk Choe, and
  Youngjoon Yoo.
\newblock {CutMix}: Regularization strategy to train strong classifiers with
  localizable features.
\newblock In {\em IEEE International Conference on Computer Vision}, pages
  6023--6032, 2019.

\bibitem{yun2020videomix}
Sangdoo Yun, Seong~Joon Oh, Byeongho Heo, Dongyoon Han, and Jinhyung Kim.
\newblock {VideoMix}: Rethinking data augmentation for video classification.
\newblock {\em arXiv preprint arXiv:2012.03457}, 2020.

\bibitem{zhang2018mixup}
Hongyi Zhang, Moustapha Cisse, Yann~N. Dauphin, and David Lopez-Paz.
\newblock {MixUp}: Beyond empirical risk minimization.
\newblock In {\em International Conference on Learning Representation}, 2018.

\bibitem{zhang2019deeper}
Zhipeng Zhang and Houwen Peng.
\newblock Deeper and wider siamese networks for real-time visual tracking.
\newblock In {\em IEEE Conference on Computer Vision and Pattern Recognition},
  pages 4591--4600, 2019.

\bibitem{Ocean_2020_ECCV}
Jianlong Fu Bing Li Weiming~Hu Zhipeng~Zhang, Houwen~Peng.
\newblock {Ocean}: Object-aware anchor-free tracking.
\newblock In {\em European Conference on Computer Vision}, pages 771--787,
  2020.

\bibitem{zhu2022srrt}
Jiawen Zhu, Xin Chen, Dong Wang, Wenda Zhao, and Huchuan Lu.
\newblock {SRRT}: Search region regulation tracking.
\newblock {\em arXiv preprint arXiv:2207.04438}, 2022.

\bibitem{zhu2018distractor}
Zheng Zhu, Qiang Wang, Bo Li, Wei Wu, Junjie Yan, and Weiming Hu.
\newblock Distractor-aware siamese networks for visual object tracking.
\newblock In {\em European Conference on Computer Vision}, pages 101--117,
  2018.

\end{thebibliography}
}
\end{document}